\documentclass[runningheads]{llncs}
\usepackage[T1]{fontenc}
\usepackage{graphicx}
\usepackage{amsmath}
\usepackage{amssymb}
\usepackage{xcolor}
\usepackage{enumitem}
\usepackage{tabularx}
\usepackage{booktabs}  
\usepackage{array}
\usepackage{hyperref}
\usepackage{subcaption}

\begin{document}
\title{Deep Spectral Methods for Unsupervised Ultrasound Image Interpretation}
%

%
\titlerunning{Deep Spectral Methods for Ultrasound}
%
\author{Oleksandra Tmenova\inst{1 * \star} \and
Yordanka Velikova\inst{1,2 *} \and
Mahdi Saleh\inst{1} \and
Nassir Navab\inst{1,2}
}%

\authorrunning{Tmenova et al.}
\institute{Computer Aided Medical Procedures, Technical University of Munich, Germany \\ 
 \and
Munich Center for Machine Learning, Munich, Germany \\
 }
\maketitle              

{\let\thefootnote\relax\footnotetext{*Shared first authorship. \(\star\) oleksandra.tmenova@tum.de}} 

\begin{abstract}
Ultrasound imaging is challenging to interpret due to non-uniform intensities, low contrast, and inherent artifacts, necessitating extensive training for non-specialists. Advanced representation with clear tissue structure separation could greatly assist clinicians in mapping underlying anatomy and distinguishing between tissue layers. Decomposing an image into semantically meaningful segments is mainly achieved using supervised segmentation algorithms. Unsupervised methods are beneficial, as acquiring large labeled datasets is difficult and costly, but despite their advantages, they still need to be explored in ultrasound. This paper proposes a novel unsupervised deep learning strategy tailored to ultrasound to obtain easily interpretable tissue separations. We integrate key concepts from unsupervised deep spectral methods, which combine spectral graph theory with deep learning methods. We utilize self-supervised transformer features for spectral clustering to generate meaningful segments based on ultrasound-specific metrics and shape and positional priors, ensuring semantic consistency across the dataset. We evaluate our unsupervised deep learning strategy on three ultrasound datasets, showcasing qualitative results across anatomical contexts without label requirements. We also conduct a comparative analysis against other clustering algorithms to demonstrate superior segmentation performance, boundary preservation, and label consistency.

\end{abstract}

\section{Introduction}

Ultrasound is commonly used in diagnostic medicine, valued for its real-time imaging capabilities and non-invasive nature, which enables regular health checkups without ionizing radiation~\cite{noble2011ultrasonic}.
However, the interpretation of ultrasound images often presents a significant challenge, necessitating specialized training or years of experience for clinicians~\cite{zu2014predicate}. 
The complexity of these images makes the apparent separation and identification of tissue structures difficult. Improved representation techniques can aid in the interpretation process, in particular in understanding the underlying anatomy and differentiation between tissue layers. 

Decomposing images into semantically meaningful regions has predominantly been tackled using supervised deep learning (DL) algorithms for segmentation~\cite{liu2019deep}. Particularly, convolutional neural networks (CNNs) and architectures like U-net~\cite{ronneberger2015u} have significantly advanced supervised ultrasound segmentation~\cite{van2019deep}. These methods excel in delineating anatomical shapes and have shown promise in automating the identification of structures across many applications, leading to improved diagnostic accuracy~\cite{liu2019deep}. 
Despite their efficacy, these methods depend on the availability of large, annotated datasets for training and are often tailored to specific anatomical structures, limiting their scalability and adaptability~\cite{siddique2021u}. 

Consequently, unsupervised learning approaches, which do not necessitate expert-reliant labeled data for training, emerge as an alternative.
Relavant works utilize graph-based methods~\cite{fz,shi2000normalized}, gradient-ascent-based algorithms~\cite{vedaldi2008quick}, SLIC-K-means-based methods~\cite{slic} and intermediate representations~\cite{cactuss,lotus}.
Such techniques have proven helpful for computer vision tasks like semantic instance segmentation and have found their application in the ultrasound domain too~\cite{ilesanmi2020multiscale,huang2020segmentation}. However, they demand careful parameter selection to avoid the loss of critical edge information~\cite{neubert2012superpixel}.
Furthermore, the resultant segments have class-agnostic labels and are primarily used as an initial step for further DL-based frameworks. 

Following traditional clustering methods, spectral clustering emerges as another unsupervised approach for identifying image segments, by constructing a similarity graph representing the relationships between data points. For spectral clustering, an affinity matrix is built, and by utilizing eigenvalues and eigenvectors, similar data points are grouped into clusters~\cite{von2007tutorial}. 
Spectral clustering excels in handling complex cluster shapes that are non-convex or consist of disjoint convex sets, making it particularly advantageous for applications where conventional clustering methods fail. 
When applied to images where the affinity matrix mirrors the adjacency matrix of a graph, spectral clustering is used to identify normalized graph cuts, which can divide images into meaningful segments without the need for predefined labels~\cite{shi2000normalized}.

Recent works leverage the strengths of self-supervised Vision Transformer (ViT) models, such as DINO~\cite{caron2021emerging}, which utilize self-distillation techniques to learn rich visual feature representations from unlabeled data. 
Those features are then applied to spectral clustering techniques to construct an affinity matrix and identify distinct segments within an image~\cite{wang2022self,wang2023tokencut,dsp,cutler}. 
In particular, deep spectral segmentation (DSS)~\cite{dsp} employs multiple eigenvectors to obtain per-image segments and introduces additional spatial and color affinities for improved consistency. It then utilizes DINO features from all dataset segments and clusters them to obtain semantic labels. 
Combining self-supervised ViT features with spectral clustering has become a powerful approach for unsupervised object discovery and segmentation.
Those unsupervised segmentation techniques have drawn attention for their ability to provide label-free representations easily adaptable for downstream tasks~\cite{balestriero2023cookbook}, making them particularly suitable for medical imaging applications where labeled data is scarce. 
\subsubsection{Contributions} 
This work introduces an unsupervised deep-learning framework specifically designed for enhancing ultrasound image analysis. Utilizing self-supervised transformer-based features, we implement spectral clustering to derive semantically meaningful segments. We incorporate ultrasound-specific metrics together with shape and geometric priors to ensure consistency across diverse anatomical contexts. This provides clear tissue structure separation without the need for labeled datasets. We validate our framework across three ultrasound datasets, showcasing its capability, and provide qualitative results that adeptly preserve the contours of the underlying anatomical structures. Our comparative analysis with other clustering algorithms underscores our method's superior segmentation accuracy, boundary preservation, and label consistency performance. The source code is publicly available at \url{https://github.com/alexaatm/UnsupervisedSegmentor4Ultrasound.git} \footnotemark[1]\footnotetext[1]{All implementation and experiments were conducted by O. Tmenova as part of her master’s thesis at TUM.}

\section{Method}\label{chapter:methodology}

\begin{figure}[h]
    \centering
    \includegraphics[width=\linewidth]{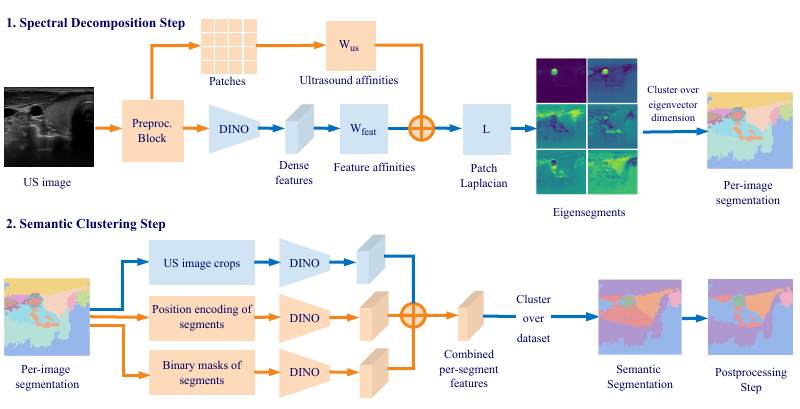}
    \caption{In our unsupervised semantic segmentation pipeline, ultrasound images undergo preprocessing and dense feature extraction to derive feature affinities. Ultrasound-specific affinities are then calculated using similarity metrics (MI, SSD) and combined with initial affinities for spectral clustering, yielding pseudo masks. Subsequently, images are cropped to focus on detected segments, and dense features alongside positional and shape priors refine clustering across the dataset. This two-step process enhances semantic consistency, transitioning from class-agnostic to more meaningful segmentations, all without relying on labels.}
    \label{fig:pipeline}
\end{figure}

Our approach builds upon the deep spectral family of unsupervised segmentation methods~\cite{wang2023tokencut,dsp,cutler}, particularly the deep spectral segmentation (DSS) for multiple-object semantic segmentation~\cite{dsp}. 
The proposed method’s architecture, shown in Figure \ref{fig:pipeline}, includes two major steps: spectral decomposition for obtaining per-image segments and clustering them into semantically consistent classes. As an addition to the duo of self-supervised transformers with classic spectral clustering ~\cite{wang2022self,dsp,cutler}, we propose several adaptations to enhance segment separation in ultrasound images.  In the first step (Figure \ref{fig:pipeline}, top), we introduce ultrasound affinities and add a preprocessing step to address the domain gap between natural and ultrasound data. In the second step (Figure \ref{fig:pipeline}, bottom), we incorporate additional shape and position priors to add extra information to the final clustering of segments.

\subsection{Spectral decomposition}
\subsubsection{Data Preprocessing}
Different from real-world images with diverse colors and distinct borders, ultrasound data is infamously challenging to analyze.
That is why US image analysis benefits from proper preprocessing~\cite{mounica2019study,che2017ultrasound}. To take this into account, we add a preprocessing block to the pipeline and explore different strategies for enhancing the image quality, including classical approaches (gaussian blurring, histogram equalization) and pretrained denoising models like MPRNet~\cite{zamir2021multi}.


\subsubsection{Affinity Matrix Construction}
Self-supervised attention-based architectures like DINO~\cite{caron2021emerging} serve as a good base for extracting rich features. Following \cite{dsp}, we use the features from the keys of the last attention layer of the pre-trained DINO.
An essential step in spectral clustering is treating image segmentation as a graph-cutting problem~\cite{shi2000normalized}. Images are represented as graphs \( G = (V, E) \) where nodes correspond to either pixels (for color affinities) or patches (for DINO affinities). Edge weights between nodes indicate their similarity. The self-correlation of DINO features provides an effective affinity matrix, enabling successful graph partitioning and meaningful image segments. Like in DSS~\cite{dsp}, the features are thresholded at 0 to exclude anti-correlations:

\begin{equation}\label{eq:feat}
    W_{\text{feat}} = \text{f} \cdot \text{f}^T \odot (\text{f} \cdot \text{f}^T > 0)
\end{equation}

Since color affinities cannot be leveraged from ultrasound greyscale data, we integrate ultrasound patch-wise affinities employing standard pixel-based metrics that proved successful in the task of both rigid and non-rigid ultrasound image registration~\cite{che2017ultrasound}. In particular, we employ two common metrics: Sum of Squared Differences (SSD) \( SSD(P_1, P_2) = \sum_{i=1}^{X} \sum_{j=1}^{Y} (P_1(i, j) - P_2(i, j))^2\)  where $X$ and $Y$ represent the dimensions of the patches $P_1$ and $P_2$, and Mutual Information (MI) \( MI(P_1, P_2) =(H(P_1) + H(P_2))/(H(P_1, P_2))\), where $H(P_1)$ and $H(P_2)$ are the entropies of the individual patches, and $H(P_1, P_2)$ is their joint entropy. To build the affinity matrix, an image is partitioned into patches of size \(k \times k\). We chose \(k\) to match the patch size of the used transformer backbone.

The dissimilarity matrix \(D_{\text{patchwise}}\) is then constructed by comparing each patch \(P_i\) to every other patch \(P_j\) using the specified distance metric \(d\) (Eq. \ref{eq:dist_matrix}).  It is then transformed into an affinity matrix using a Gaussian kernel (Eq. \ref{eq:aff_matrix}). 
\begin{equation}\label{eq:dist_matrix}
    D_{\text{patchwise}}(P_i, P_j) = d(P_i, P_j) = \begin{cases} \text{SSD}(P_i, P_j), & \text{if } d = \text{SSD} \\ 
        1 - \text{MI}(P_i, P_j), & \text{if } d = \text{MI} \end{cases} ,
\end{equation}
\begin{equation}\label{eq:aff_matrix}
    W_{\text{patchwise}} = \exp\left(- \delta \cdot D_{\text{patchwise}}\right)
\end{equation}
Additionally, we explore position-based affinities using linear interpolation from \(0\) to \(1\) for the \(N_{height}\) and \(N_{width}\), 
where \(N_{height} = H // k \), \(N_{width} = W // k \), where \( k\) is the size of a patch, which 
results in patch feature vectors $\psi(u) = (x_{\text{pos}}, y_{\text{pos}})$, which are then used to construct a positional affinity matrix (Eq. \ref{eq:pos_aff}).
\begin{equation}\label{eq:pos_aff}
    W_{\mathrm{pos}}(P_i, P_j) =
     \begin{cases}
        1 - \|\psi(P_i) - \psi(P_j)\|, & \mathrm{if \;} P_i \in \mathrm{KNN}_{\psi}(P_j), \\
        0,  & \mathrm{otherwise}, 
    \end{cases} 
\end{equation}
where \(P_i \in \mathrm{KNN}_{\psi}(P_j)\) are the \(k\)-nearest neighbors of patch \(P_j\) under the SSD distance of feature vectors \(\psi\).
Finally, we linearly combine DINO, ultrasound, and positional affinities, controlled by coefficients \(C_{feat}, C_{mi}, C_{pos}\), to obtain the final affinity matrix needed for spectral clustering (Eq. \ref{eq:aff_comb}).
\begin{equation}\label{eq:aff_comb}
    W_{\text{comb}} = W_{\text{feat}} + C_{\text{ssd}} \cdot W_{\text{ssd}} + C_{\text{mi}} \cdot W_{\text{mi}} + C_{\text{pos}} \cdot W_{\text{pos}},
\end{equation}

\subsubsection{Spectral Clustering}
From the obtained affinity matrix \(W_{comb}\), its Laplacian matrix is calculated (Eq. \ref{eq:lapl}). Then the objective function for spectral clustering can be expressed using the graph Laplacian:
\(\min \text{Tr}(E^\top L E) \text{ s.t. } E^\top E = I\), 
where \(\text{Tr}\) denotes the trace norm of a matrix, and \(E=\{a_{ij}\}\) is a matrix whose rows represent the low-dimensional embedding of the original data points. 
\begin{equation}\label{eq:lapl}
    L = D^{-1/2}(D - W)D^{-1/2}, \text{ where } D \text{ has values } d_{ii} = \sum_{j} a_{ij} \quad \text{for all } i
\end{equation}


The Laplacian matrix is decomposed into eigensegments, \(e_0, \ldots, e_{n-1}\), where only positive eigenvectors (\(e > 0\)) are used as per-image segments. K-means clustering is then applied to obtain these segments, following the approach in DSS~\cite{dsp}. We refer to this step as Oversegmentation, with the number of eigensegments set to 15.

\subsection{Semantic Clustering} 

In the second clustering step, bounding boxes of segments are calculated to extract per-segment features, which are then clustered using K-means \cite{dsp}. We refine this process for ultrasound data and optimize the segment feature extraction step by addressing the challenge of textural similarity in different anatomical areas.
To achieve this, we employ a dual embedding strategy that enhances the differences between features of segments, such as vessels and features of other areas with similar textures, while minimizing the overall segment count. We construct a mask embedding to capture shape features via binary masks and positional embedding to encode spatial locations of segments. This streamlined approach ensures that segments are grouped not only by similar features but also by shape and position, resulting in  better  spatial and structural consistency across ultrasound sweeps.
The resulting feature vectors from the image crop \( \text{f}_{\text{image}} = \phi(\text{s}_{\text{crop}}) \), from the mask \( \text{f}_{\text{mask}} = \phi(\text{s}_{\text{mask}}) \), and from the position encoding \( \text{f}_{\text{pos}} = \phi(\text{s}_{\text{pos}}) \) are then linearly combined before clustering.
\subsubsection{Postprocessing}
Results obtained after the two clustering steps can already serve as a coarse segmentation. However, for sharper boundaries, we include additional postprocessing. We upscale and apply CRF~\cite{crf}, as also commonly done in other segmentation pipelines~\cite{dsp,cutler,stego}.

\begin{figure}[h]
    \centering
    \includegraphics[width=0.95 \linewidth]{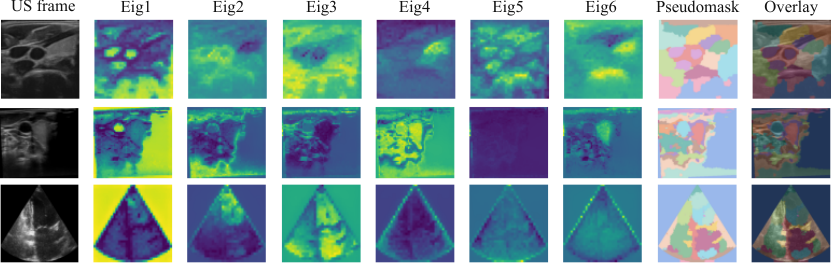}
    \caption{Combining ultrasound-based affinities and deep features leads to meaningful image separation. }
    \label{fig:teaser}
\end{figure}

\section{Experimental Setup}
\textbf{CCA} Common Carotid Artery dataset consists of ultrasound images from four different machines from 24 adults with single labels of the carotid artery~\cite{CCA_data,CCA_data2}, which was sampled to remove repetitive slices, totaling 349 images for testing. 

\noindent \textbf{Thyroid} dataset contains annotated 3D ultrasound images of the thyroid~\cite{thyData}. It includes scans from 28 healthy volunteers using a Siemens Acuson NX-3 US machine with a VF12-4 probe. The 3D ultrasound scans were post-processed to remove empty labels, extract the 2D slices with corresponding labels, and remove repetitive slices, in total 634 images. 

\noindent \textbf{CAMUS} dataset includes 400 cardiac patient images for training and 50 for test~\cite{camusData}. For our evaluation, we used end-systole (ES) and end-diastole (ED) images from the test set - in total 500 validation images (5 for each ES and ED) from 50 patients with 3 labels from manual expert annotations of the left ventricle endocardium, the myocardium and the left atrium.

\noindent \textbf{Evaluation Methodology}
The evaluation methodology includes two main aspects: per-image mask evaluation (step I) and semantic 
evaluation post-clustering (step II). For step I, we assess the quality of individual segments before clustering
using DICE score.
For step II, we evaluate the segments obtained after semantic clustering. Ground truth and pseudo-labels are matched using Hungarian matching or majority vote \cite{kirillov2019panoptic,dsp}, and only matched masks are evaluated. To assess semantic consistency of label mappings, we identify the most prevalent pseudo-label class assigned to each ground truth class across the entire dataset. 
The label consistency (LC) metric is then computed as the percentage of times the final pseudo-label class has been consistently assigned to a particular ground truth class across the entire dataset. 
We use DSS \cite{dsp} as a baseline for comparisons, which aligns with our goal of multi-class segmentation. TokenCut \cite{wang2023tokencut} and CutLER \cite{cutler}, while conceptually similar, focus on single-object and instance segmentation, respectively, making direct comparisons challenging. Therefore, we focus on zero-shot unsupervised methods that segment images
without prior training: SLIC \cite{slic} and FZ \cite{fz}, baselines for superpixel evaluation, including ultrasound \cite{usSuperpixel}. We report superpixel metrics such as Boundary Recall (BR) and Undersegmentation Error (UE) \cite{neubert2012superpixel}, setting the distance parameter $d$ to 3 to accommodate the imprecise boundaries in ultrasound images.

\begin{table}[b]
\centering
\caption{Comparison with baseline method STEP 1. *CRF postprocessing}
\label{tab:eval_baseline}
\resizebox{0.7 \textwidth}{!}{%
\begin{tabular}{l|c|c|c|c}
\toprule
\textbf{Method} & \textbf{N seg} & \textbf{Carotid} & \textbf{Thyroid} & \textbf{Cardiac} \\
 &  & \textbf{DICE, std} & \textbf{DICE, std} & \textbf{DICE, std} \\ 
 \midrule
DSS baseline * & 15 & 32.33 $\pm$ 11.38 & 43.75 $\pm$ 9.87 & 36.98 $\pm$ 8.49 \\
\midrule
Ours\(_{proc}\)*      & 15 & 56.31 $\pm$ 12.89 & 62.45 $\pm$ 10.91 & 42.13 $\pm$ 6.78 \\
Ours\(_{Aff}\)*        & 15 & \textbf{63.72 $\pm$ 14.31} & \textbf{62.52 $\pm$ 8.62} & 40.44 $\pm$ 9.07 \\
Ours\(_{comb}\)*        & 15 & 46.21 $\pm$ 8.43 & 61.43 $\pm$ 10.03 & \textbf{45.32} $\pm$ \textbf{9.16} \\ 
\bottomrule
\end{tabular}%
}
\end{table}

\begin{table}[h]
\centering
\caption{Evaluation on downstream task STEP II}
\label{tab:eval_baseline_step2}
\resizebox{0.9\textwidth}{!}{%
\begin{tabular}{l|cc|cc|cc}
\toprule
\textbf{Method} & \multicolumn{2}{c|}{\textbf{Carotid}} & \multicolumn{2}{c|}{\textbf{Thyroid1}} & \multicolumn{2}{|c}{\textbf{Cardiac}} \\
 & \textbf{DICE, std} & \textbf{LC, std} & \textbf{DICE, std} & \textbf{LC, std} & \textbf{DICE, std} & \textbf{LC, std} \\ \midrule
DSS baseline & $39.24 \pm 9.1$ & $\textbf{70.25} \pm \textbf{16.0}$ & $39.57 \pm 8.6$ & $47.77 \pm 15.4$ & $26.12 \pm 7.7$ & $78.33 \pm 16.5$ \\
Ours\(_{proc}\) + DSS\(_{step2}\) & $32.25 \pm 6.7$ & $47.69 \pm 14.8$ & $50.44 \pm 7.3$ & $67.61 \pm 9.8$ & $25.38 \pm 7.3$ & $77.06 \pm 10.8$ \\
Ours\(_{Aff}\) +  DSS\(_{step2}\) & $42.56 \pm 10.0$ & $55.72 \pm 10.8$ & $54.95 \pm 10.4$ & $\textbf{71.99} \pm \textbf{3.5}$ & $\textbf{37.53} \pm \textbf{6.5}$ & $85.00 \pm 10.8$ \\
Ours\(_{comb}\) +  DSS\(_{step2}\) & $30.50 \pm 7.3$ & $55.50 \pm 13.3$ & $\textbf{59.86} \pm \textbf{8.7}$ & $59.79 \pm 8.9$ & $29.25 \pm 10.6$ & $87.38 \pm 11.3$ \\
Ours\(_{proc}\) +  Ours\(_{step2}\) & $32.32 \pm 6.9$ & $52.50 \pm 14.6$ & $50.26 \pm 17.9$ & $67.01 \pm 10.1$ & $26.82 \pm 11.3$ & $\textbf{93.75} \pm \textbf{8.2}$ \\
Ours\(_{Aff}\) +  Ours\(_{step2}\) & $\textbf{44.98} \pm \textbf{14.5}$ & $52.18 \pm 15.9$ & $47.62 \pm 11.1$ & $63.77 \pm 14.0$ & $30.92 \pm 9.4$ & $81.56 \pm 11.9$ \\
Ours\(_{comb}\) +  Ours\(_{step2}\) & $19.30 \pm 11.2$ & $45.14 \pm 5.1$ & $52.90 \pm 10.7$ & $74.63 \pm 8.9$ & $28.11 \pm 13.3$ & $85.71 \pm 9.2$ \\ \bottomrule
\end{tabular}%
}
\end{table}

\section{Results and Discussion}\label{chapter:dicussion}

In Tables \ref{tab:eval_baseline} and \ref{tab:eval_baseline_step2}, we compare the performance of our proposed method against the DSS baseline \cite{dsp} with added preprocessing ($ours_{preproc}$), affinities ($ours_{aff}$), and their combined effect ($ours_{comb}$) in terms of the DICE score. In Table \ref{tab:eval_baseline_step2} we additionally assess their effect together with positional and mask priors in the semantic clustering step and evaluate label consistency (LC).
Our proposed methods ($ours_{preproc}$,  $ours_{aff}$ $ours_{comb}$) show improvements in segmentation quality compared to the baseline method \cite{dsp} across all three datasets. Specifically, $ours_{aff}$ consistently achieves the highest DICE scores of $63.72\pm14.31$ for the Carotid dataset, $62.52\pm8.62$ for the Thyroid dataset, and $45.32\pm9.16$ for Cardiac dataset. The improvement in segmentation quality suggests that the preprocessing steps and additional affinities positively enhance segmentation performance. 
Figure \ref{fig:teaser} depicts eigensegments obtained from combining ultrasound MI, SSD, and positional affinities (with coefficients $1.0$, $1.0$, and $0.1$, respectively) and the resulting pseudo mask. It can be observed how different eigensegments capture distinct areas from the original image, for example, the vessels in the carotid image (top, Eig1), the thyroid lobe (middle, Eig4), or the heart chamber (bottom, Eig2), which then get assigned a distinct label.
In Table \ref{tab:eval_baseline_step2}, we observe the positive effects of preprocessing, affinities, and shape priors on semantic clustering, with DICE scores of  $44.98 \pm14.5$,  $59.86 \pm8.7$ and $37.53 \pm6.5$, consistently outperforming the baseline. However, there is a trade-off between mask quality and label consistency: methods that preserve finer details (higher DICE) result in more complex and varied segment shapes, making it harder to achieve consistent clustering labels (lower label consistency) across similar structures.

Finally, we compare the best results from Tables \ref{tab:eval_baseline} and \ref{tab:eval_baseline_step2} to SLIC \cite{slic} and Felzenszwalb \cite{fz}, common baselines for superpixel evaluation. The results are reported in Tables \ref{tab:eval_1} and \ref{tab:eval_2}.
In Table \ref{tab:eval_1}, we observe that the performance of 'deep spectral segments' is on par with SLIC and Felzenszwalb, exhibiting a lower UE of $0.0158$ and $0.2125$ for the Carotid and Cardiac datasets, respectively, and a higher BR of $0.677$ for the Thyroid dataset. SLIC has a better BR for the other two datasets, which can be explained by the fact that it is not possible to enforce a specific number of segments for fair comparisons, making the SLIC images being even more oversegmented, leading to higher BR.
In Table \ref{tab:eval_2}, we compare our eigensegments with SLIC and Felzenszwalb for a downstream task of semantic segmentation and observe that our method has both lower UE and higher BR for two out of three of our datasets.

\begin{table}[h]
\centering
\makebox[0pt][c]{\parbox{\textwidth}{%
    \begin{minipage}[t]{0.5\textwidth}\centering
        \caption{Comparison with other methods - UE and BR of Step I masks}
        \label{tab:eval_1}
        \resizebox{0.8\textwidth}{!}{%
        \begin{tabular}{l|cc|cc|cc}
        \toprule
        \textbf{Method} & \multicolumn{2}{c|}{\textbf{Carotid}} & \multicolumn{2}{c|}{\textbf{Thyroid}} & \multicolumn{2}{c}{\textbf{Cardiac}} \\
         & \textbf{UE} & \textbf{BR} & \textbf{UE} & \textbf{BR} & \textbf{UE} & \textbf{BR} \\ \midrule
        SLIC & 0.018 & \textbf{0.907} & \textbf{0.035} & 0.589 & 0.224 & \textbf{0.492} \\
        Fz & 0.026 & 0.578 & 0.035 & 0.475 & 0.302 & 0.434 \\
        Ours best & \textbf{0.016} & 0.679 & 0.051 & \textbf{0.677} & \textbf{0.213} & 0.479 \\
        \bottomrule
        \end{tabular}%
        }
    \end{minipage}%
    \hfill
    \begin{minipage}[t]{0.5\textwidth}\centering
        \caption{Comparison with other methods - UE and BR of Step II masks}
        \label{tab:eval_2}
        \resizebox{0.9\textwidth}{!}{%
        \begin{tabular}{l|cc|cc|cc}
        \toprule
        \textbf{Method} & \multicolumn{2}{c|}{\textbf{Carotid}} & \multicolumn{2}{c|}{\textbf{Thyroid}} & \multicolumn{2}{c}{\textbf{Cardiac}} \\
         & \textbf{UE} & \textbf{BR} & \textbf{UE} & \textbf{BR} & \textbf{UE} & \textbf{BR} \\ \midrule
        SLIC + DSS\(_{step2}\) & 0.046 & 0.352 & 0.126 & 0.287 & \textbf{0.139} & \textbf{0.649} \\
        Fz + DSS\(_{step2}\) & 0.046 & 0.335 & 0.111 & 0.314 & 0.238 & 0.339 \\
        Ours best & \textbf{0.030} & \textbf{0.433} & \textbf{0.042} & \textbf{0.589} & 0.275 & 0.379 \\
        \bottomrule
        \end{tabular}%
        }
    \end{minipage}%
}}
\end{table}

Our analysis reveals that masks derived from spectral decomposition (step I)  outperform the baseline by a large margin, showing the benefits of ultrasound-based affinities.
At the same time, final segmentation masks (step II) fall behind in DICE scores, highlighting a quality gap and the need for ensuring semantic consistency.
Although mask and position embeddings have marginally improved segmentation performance, challenges such as segment merging persist, indicating the need for further exploration into feature space enhancement and self-training techniques.

\section{Conclusions}

We present an adapted deep spectral segmentation method tailored for B-mode ultrasound data, utilizing self-supervised transformers to create affinity graphs for segment extraction. We integrate image preprocessing and leverage ultrasound-specific patchwise affinities in spectral clustering to mitigate semantic inconsistencies through mask and positional embeddings. Through extensive ablation studies, we underscore the efficacy of our approach. Our results highlight the significant potential of deep spectral methods for unsupervised ultrasound segmentation and suggest a promising direction for future investigations. 

\section*{Acknowledgements}
We would like to thank ImFusion for support and collaboration within the ForNero Project funded by BFS, AZ-1592-23.


%
%
%
\bibliographystyle{splncs04}
\bibliography{bibliography}

\appendix
\leavevmode\newline
%
\def\etal{~\textit{et.al.}}
%
          
\newpage
\setcounter{figure}{0}  
\section*{Supplementary Material} \leavevmode
\subsection*{Additional Qualitative Results}

\begin{figure}[h]
    \centering
\includegraphics[width=\textwidth]{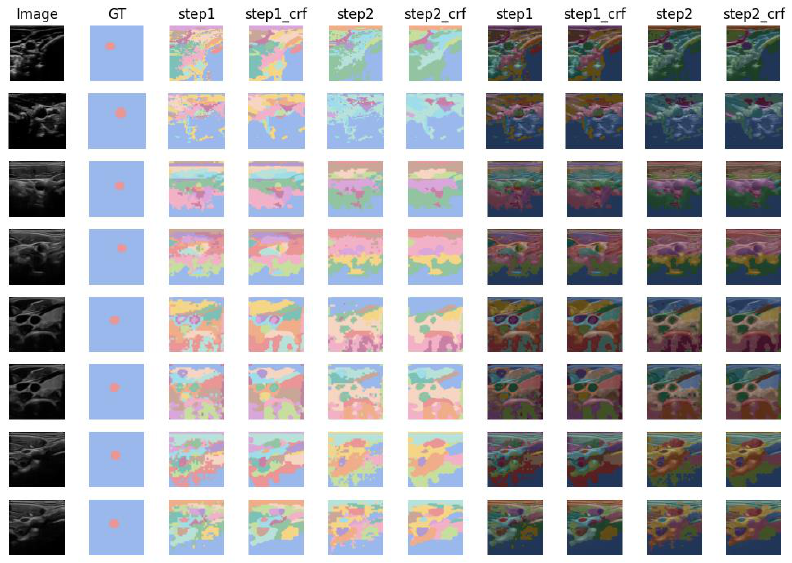} 
\caption{Additional results for Carotid dataset (Ours\(_{proc}\)).}
\end{figure}

\begin{figure}[h]
    \begin{subfigure}{\textwidth}
        \centering
        \includegraphics[width=\textwidth]{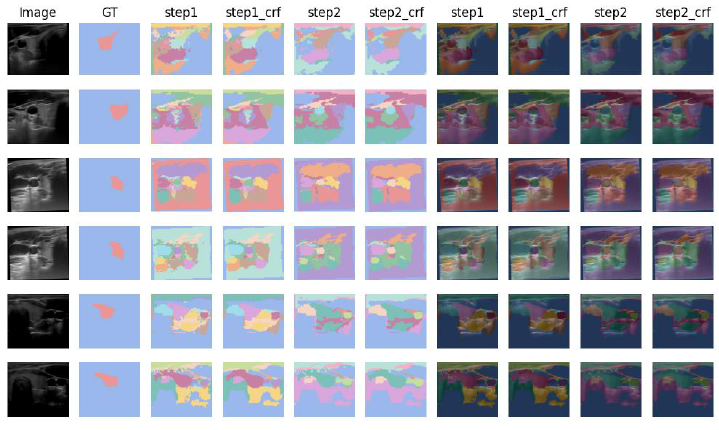}
        \caption{Additional results for Thyroid dataset (Ours\(_{proc}\)).}
    \end{subfigure}
    \vspace{3em} 
    
    \begin{subfigure}{\textwidth}
        \centering
        \includegraphics[width=\textwidth]{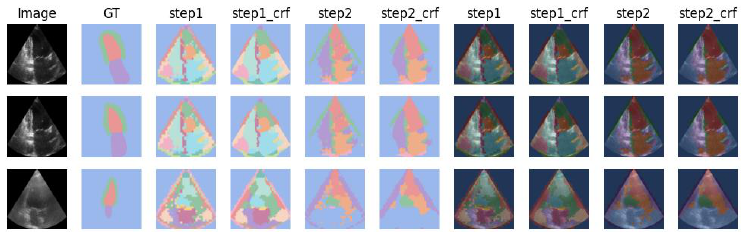}
        \caption{Additional results for CAMUS dataset.}
    \end{subfigure}

\end{figure}

\begin{figure}[h]
    \centering
    
    \begin{subfigure}{\textwidth}
        \centering
        \includegraphics[width=0.9\textwidth]{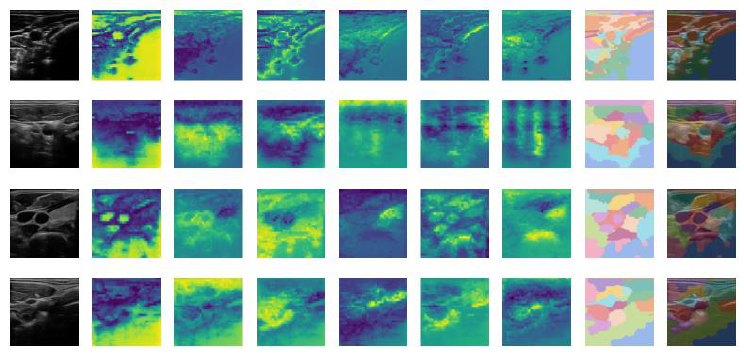}
        \caption{Eigenvectors from ultrasound affinities on Carotid dataset.}
    \end{subfigure}
    \vspace{2em}
    
    \begin{subfigure}{\textwidth}
        \centering
        \includegraphics[width=0.9\textwidth]{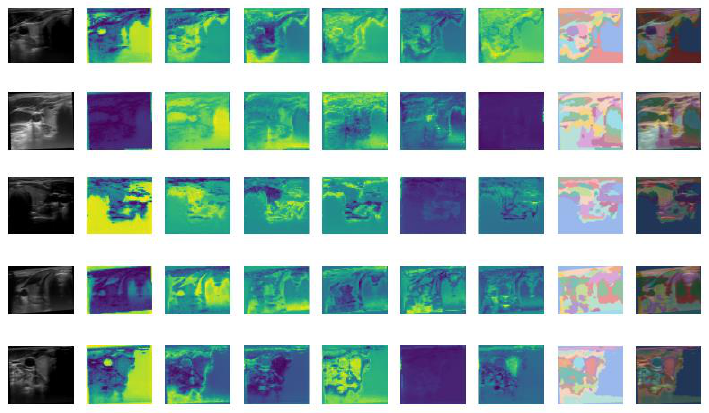}
        \caption{Eigenvectors from ultrasound affinities on Thyroid dataset.}
    \end{subfigure}
    
    \vspace{2em}
    
    \begin{subfigure}{\textwidth}
        \centering
        \includegraphics[width=0.9\textwidth]{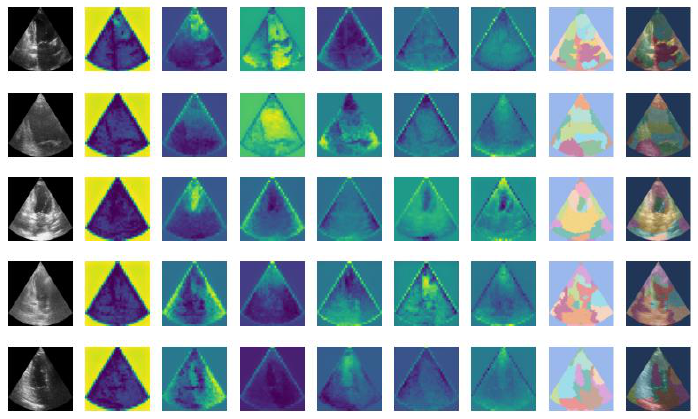}
        \caption{Eigenvectors from ultrasound affinities on CAMUS dataset.}
    \end{subfigure}

\end{figure}


\end{document}